\documentclass[conference]{IEEEtran}
\pdfoutput=1
\usepackage[hyphens]{url}
\usepackage{hyperref}
\hypersetup{colorlinks,allcolors=blue}
\usepackage[backend=biber, style=ieee]{biblatex}

\IEEEoverridecommandlockouts
\usepackage{amsmath,amssymb,amsfonts}
\usepackage{algorithmic}
\usepackage{graphicx}
\usepackage{textcomp}
\usepackage[table,xcdraw]{xcolor}
\usepackage{balance}
\usepackage{mathtools}  
\usepackage{orcidlink}  

\def\BibTeX{{\rm B\kern-.05em{\sc i\kern-.025em b}\kern-.08em
    T\kern-.1667em\lower.7ex\hbox{E}\kern-.125emX}}

\usepackage{geometry}
\begin{document}

\title{Using LLMs to Establish Implicit User Sentiment of Software Desirability\textsuperscript{*}
  \thanks{\textsuperscript{*}The authors wish to thank Creighton University and Dakota State University for their generous financial support of this research.}
}

\author{Sherri Weitl-Harms\,\orcidlink{0000-0002-3653-2928}\textsuperscript{†}, John D. Hastings\,\orcidlink{0000-0003-0871-3622}\textsuperscript{‡}, Jonah Lum\,\orcidlink{0009-0004-2428-6047}\textsuperscript{§}


  \thanks{\textsuperscript{†}Department of Computer Science, Design \& Journalism (CSDJ), Creighton University, Omaha, NE, USA. Email: sherriweitlharms@creighton.edu}
\thanks{\textsuperscript{‡}The Beacom College of Computer \& Cyber Sciences, Dakota State University, Madison, SD, USA. Email: john.hastings@dsu.edu}
\thanks{\textsuperscript{§}Department of Computer Science, Design \& Journalism (CSDJ), Creighton University, Omaha, NE, USA. Email: jonahlum@creighton.edu}

}

\maketitle

\begin{abstract}
This study explores the use of  LLMs for providing quantitative zero-shot sentiment analysis of implicit software desirability, addressing a critical challenge in product evaluation where traditional review scores, though convenient, fail to capture the richness of qualitative user feedback. Innovations include establishing a  method that 1) works with qualitative user experience data without the need for explicit review scores, 2) focuses on implicit user satisfaction, and 3) provides scaled numerical sentiment analysis, offering a more nuanced understanding of user sentiment, instead of simply classifying sentiment as positive, neutral, or negative.

Data is collected using the Microsoft Product Desirability Toolkit (PDT), a well-known qualitative user experience analysis tool. For initial exploration, the PDT metric was given to users of two software systems. PDT data was fed through several LLMs (Claude Sonnet 3 and 3.5, GPT4, and GPT4o) and through a leading transfer learning technique, Twitter-Roberta-Base-Sentiment, and Vader, a leading sentiment analysis tool. Each system was asked to evaluate the data in two ways, by looking at the sentiment expressed in the PDT word/explanation pairs; and by looking at the sentiment expressed by the users in their grouped selection of five words and explanations, as a whole. Numerical analysis is used to provide insights into the magnitude of sentiment to drive high quality decisions regarding product desirability. Each LLM is asked to provide its confidence (low, medium, high) in its sentiment score, along with an explanation of its score.

All LLMs tested were able to statistically detect user sentiment from the users' grouped data, whereas TRBS and Vader were not. The confidence and explanation of confidence provided by the LLMs assisted in understanding user sentiment. This study adds deeper understanding of evaluating user experiences, toward the goal of creating a universal tool that quantifies implicit sentiment.

\end{abstract}


\section{Introduction}\label{introduction}
In product development, understanding implicit user sentiment is crucial for creating products that truly appeal to their intended audience. It is also important for improving and marketing products. Sentiment analysis, a field dedicated to automatically extracting emotional reactions expressed by users \cite{Zhang2018}, holds the potential to provide deeper insights into user perceptions, satisfaction, and overall product desirability. 

Recent advancements in natural language processing (NLP), particularly the development of large language models (LLMs), have opened new possibilities for sentiment analysis, particular for qualitative data. These models have demonstrated capabilities in various tasks, including zero-shot sentiment classification \cite{wang2023chatgpt}.

Often, user reviews or expressions on social media are used  as a basis for sentiment analysis.  Understanding implicit sentiment, a common linguistic phenomenon, where accurate judgment often requires common sense or domain knowledge, is challenging \cite{wang2023chatgpt}. Quantifying implicit user sentiment remains a significant challenge, particularly when explicit user ratings or reviews are unavailable \cite{zhou2021implicit}.

To address situations in which sentiment data is lacking, tools such as surveys and the Microsoft Product Desirability Toolkit (PDT) can be used to evaluate user experiences. The PDT~\cite{Benedek,benedek2002a} is recognized as a valuable qualitative tool for evaluating user experience and satisfaction. 

This research aims to bridge the gap between qualitative sentiment data and quantitative analysis by applying recent LLMs to PDT data. The goal is to derive meaningful quantitative insights from qualitative user responses. In addition, by comparing LLMs with existing approaches, the study seeks to identify the most effective methods for quantifying implicit user sentiment in software desirability evaluations. The following research questions guide the study:

\begin{enumerate}
   \item[1)] How do various LLMs compare to known tools in providing sentiment analysis scores of PDT data?
     \item[2)] Is there a particular algorithm that outperforms others and could be used as the basis for the development of an implicit user sentiment analysis tool?
\end{enumerate}

The sections that follow provide background information on the PDT and sentiment analysis techniques, detail the methodology, present results, discuss their implications, and outline directions for future work.

\section{Background}\label{background}

\subsection{Product Desirability Toolkit}

The PDT is a well-known qualitative analysis tool used to evaluate user experience and satisfaction with products, such as software \cite{Barnum2010, Barnum, Booth2013EndUserEO, Hastings, Li2014, Tullis, Veral, Weitl}. It aims to ``understand the illusive, intangible aspect of desirability resulting from a user’s experience with a product'' \cite{Barnum}.

The PDT asks users to select five adjectives from a given set that best describe  their feelings about the experience along with providing an optional explanation of their word choices. By gathering this group of word/explanation pairs, the approach is designed to capture rich, qualitative data about user experiences and perceptions.

The advantages of using the PDT are ``1) it aims to avoid a bias toward the positive found in typical questionnaires (e.g., it has been found that if a respondent thinks that a survey intends to assess the quality of a product, they are likely to provide more positive answers about quality) and 2) it is able to more effectively uncover constructive negative criticisms in the guided interview'' \cite{Hastings}.

The PDT is described as the closest tool that uses ``psychometric theory to create a user experience (UX)-relevant measure of product or service desirability'' \cite{Sauro}. The design of the PDT prompts users to tell a revealing story of their experience as users comment on their word choice \cite{Barnum2010} and provides a rich set of qualitative data related to the user's implicit desirability of the product in question. 

While the PDT is a great qualitative tool, it is a poor quantitative tool by itself \cite{Sauro}. The PDT provides a way to triangulate findings from other feedback mechanisms, with potential to produce more meaningful and substantive results of user experiences \cite{Barnum2010}, but it necessary to think of improvements on the original method \cite{Veral}. The PDT text from respondent word/comments groupings are ripe for sentiment analysis.

\subsection{Sentiment Analysis}

Sentiment analysis, a field dedicated to automatically extracting emotional reactions expressed by users \cite{Zhang2018}, holds broad applications, from influencing policy decisions, refining product development \cite{Bharadwaj}, managing brand reputation, and handling crisis communication \cite{kaur,Tsai, Sharma}. Businesses derive value from sentiment analysis for market research, understanding user perception, and customer satisfaction \cite{Tsai,Venkata}. 

Sentiment analysis employs NLP, statistical methods, and machine learning algorithms to identify trends and patterns among attitudes, emotions, and opinions expressed in text, subsequently classifying them into categories or sentiment scores \cite{Tsai,Medhat,Klongdee, Silva}. Quantifying implicit user sentiment is challenging, but important for understanding user experiences.

Traditional sentiment analysis techniques often rely on lexicon-based approaches or machine learning models trained on labeled data. However, these methods can struggle with context-dependent sentiments and implicit expressions of user feelings \cite{dang2020sentiment}, which are common in PDT responses.

\subsection{LLMs in Sentiment Analysis}

LLMs have shown promising results in various NLP tasks, including their ability to guage sentiment effectively \cite{Tsai, Susnjak, Krugmann, Akihito, Kazi}. LLMs, such as GPT models, are trained on vast amounts of text data and can generate human-like text based on input prompts. As of late 2023, GPT (gpt-3.5-turbo-0301) demonstrated impressive zero-shot capabilities in sentiment classification tasks, and could serve as a universal and well-behaved sentiment analyzer \cite{wang2023chatgpt}. Generative AI is pioneered in zero-shot content analysis, such as automated textual analysis \cite{Krugmann}. However, previous work found that LLMs generally perform poorly on implicit sentiment analysis and domain-specific training was needed to improve performance \cite{wang2023chatgpt}. The application of LLMs to implicit sentiment analysis offers new possibilities for understanding user experiences and product desirability. Their ability to understand context and capture nuanced and implicit sentiments could potentially overcome some of the limitations of traditional sentiment analysis techniques when applied to PDT data.

\section{Methods}
This research experiment focuses on sentiment analysis of PDT survey datasets, each containing five words and explanations from respondents. The goal is to evaluate the effectiveness of various sentiment analysis technologies, particularly LLMs, on this data. Unlike traditional methods that classify sentiment into three categories (positive, neutral, and negative) as seen in \cite{Krugmann}, this study employs a scaled numerical sentiment analysis ranging from 0 to 1 (with 0 being the most negative and 1 the most positive). While numerical analysis is more challenging, it offers significant advantages by providing more detail and insights into the magnitude of sentiment, which can help decision makers make better decisions regarding product desirability.

\subsection{Data Collection}

The data for this study  was previously collected \cite{Weitl} using PDT from users of ZORQ \cite{Hastings2022}, a gamification framework utilized in undergraduate computer science education, and CARMA \cite{Hastings}, a grasshopper infestation rangeland management system. The PDT data collected utilizing the same set of 55 words as \citeauthor*{Barnum2010} \cite{Barnum2010} and shown in the original article \cite{Benedek}. PDT Respondent Term Groupings (PRTG) were collected (56 for CARMA and 50 for ZORQ), where each PRTG is the group of five word/explanation pairs for one respondent.\footnote{Due to space limitations, detailed results are shown for only the ZORQ dataset, but the same process was used for both sets.}

Prior to analysis, the raw survey data was processed to create a more concise dataset in which each row consisted of a word choice and its corresponding explanation. The data were cleaned to remove any inconsistencies or errors. This included checking for and addressing issues such as missing values, mismatched quotes, and non-text characters.

\subsection{Data Labeling}
To establish a gold standard for evaluation, the authors performed manual data labeling. Many studies use LLMs such as GPT4 for data labeling \cite{He}. However, a baseline of manually labeled test data is necessary to understand the accuracy of LLM-based labeling. In-house manual data labeling secures the highest quality labeling possible and is generally considered the gold standard by data scientists and engineers \cite{labelData}. It is a common method, especially for small data sets that require expertise that cannot be easily crowd-sourced \cite{labelData} and to provide a baseline of labeled test data \cite{He, wang2023chatgpt, Mukhin}. \citeauthor*{He} \cite{He} found that GPT4 itself had higher accuracy than crowd-sourcing (particularly Amazon Mechanical Turk (MTurk) workers pipeline \cite{Murk}), highlighting the unreliable quality in crowd-sourced labels; further supporting the necessity for manual in-house labeling in this study. 

The three authors, with over 60 years of combined software development experience, have the necessary expertise to manually annotated the dataset to create a gold-standard to measure the LLMs. The annotation was completed in two ways: 1) by assigning an overall sentiment score for each user's PRTG, and 2) by scoring each individual term and explanation pair. The inter-annotator agreement (Pearson's coefficients) between the three annotators for the ZORQ PRTGs was (0.92, 0.96, and 0.96)  and (0.88, 0.89, 0.89) for the word/explanation pairs. 

\begin{figure}[htbp]
\centerline{\includegraphics[scale=0.55]
{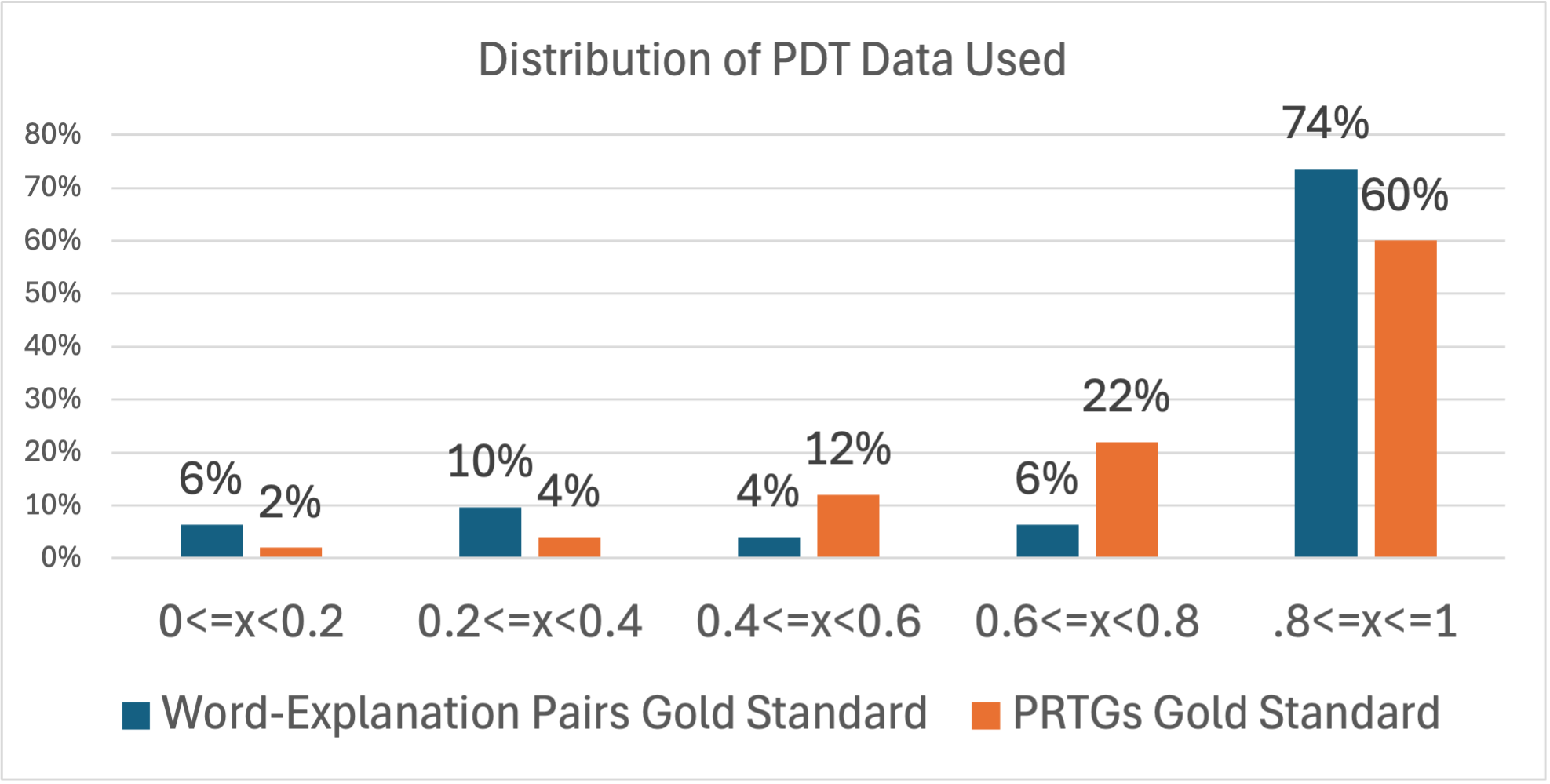}}
 \caption{PDT Data Sentiment Distribution}
 \label{DataDistribution}
 \end{figure}

The authors expect product usability sentiment to likely be skewed negatively or positively, rather than following a normal distribution. Based on the manual data labeling, the ZORQ PDT dataset had an average overall gold-standard sentiment rating of 0.76, for both the PRTGs and the word/explanation pairs, with standard deviations of 0.26 for the word/explanation pairs, and 0.19 for the PRTGs; showing that the data is skewed in the positive direction, and not normally distributed. The distribution of the data grouped into five bins, representing strongly negative to strongly positive, is shown in Fig. \ref{DataDistribution}.

\begin{table*}[!htbp]
\caption{LLM prompts}
\label{tab:llm_prompts}
  \centering
\tiny
\begin{tabular}{|p{0.09\textwidth}|p{0.85\textwidth}|}
\hline
\textbf{LLM Test} & \textbf{Prompt} \\ \hline

Claude3-Avg5 & A group of people were given a survey to assess an experience. Each respondent provided one word to describe their experience and an optional explanation for their choice. The attached spreadsheet contains the responses, one word response per row along with the explanation. In CSV form, provide a sentiment score between 0.00-1.00 (inclusive) for each of the chosen words based on your understanding of the word (with 1.00 being the most positive), along with an adjusted sentiment score for the word based on the context provided by the corresponding explanation. If no explanation is given for a word, the base and adjusted scores should be the same. Each row of output should contain the word, the base score, the adjusted score, and the explanation. Sentiment scores should be to two decimal places. It is very important that you output the results for all 250 responses. \\ \hline

Claude3.5-Respondent & The attached file contains survey responses from 50 respondents. Each respondent selected up to five words to describe their experience, and for each word, optionally provided an explanation for the choice. Please produce an overall sentiment score to two decimal places between 0.00-1.00 (inclusive) for each respondent based on your understanding of language. For each sentiment score, please also provide your confidence in the score (low, medium, high) and a detailed contextual explanation for the sentiment score that a non-technical person can understand. \\ \hline

GPT4-Avg5 \& GPT4o-Avg5 & The following lines contain word choices, which are sometimes followed by explanations for the choice. For these lines, provide a sentiment analysis score for each word between 0.00-1.00 (to two decimal places), and then an adjusted score for the word based on the explanation. Include your confidence in the accuracy of that score (low, medium, high). Additionally, provide a carefully crafted contextual explanation for the sentiment score that is related to the meaning of the text. Please provide your response in a text-based csv format, with columns for the word, original score, adjusted score, confidence, and explanation. Please do not provide any other response aside from the csv formatted data. Only provide one response per line: \\ \hline

GPT4-Respondent \& GPT4o-Respondent & Each of the following lines contains a word choice, sometimes followed by an explanation for the choice. Based on this data, please provide a singular sentiment analysis score for all of the words between 0.00-1.00 (to two decimal places), and then a singular adjusted score for all of the words based on all of the explanations. In the event that explanations are not provided with the word choices, please make the adjusted score the same as the original. Include your confidence in the accuracy of that score (low, medium, high). Additionally, provide a carefully crafted contextual explanation for the sentiment score that is related to the meaning of the texts. Please provide your response in a text-based csv format, with columns for the original score, adjusted score, confidence, and explanation. Please do not provide any other response aside from the csv formatted data. Please do not evaluate each line individually, evaluate all of the lines as a whole: \\ \hline
\end{tabular}
\end{table*}

\subsection{Sentiment Analysis Tools and Methods} 

This research explores sentiment analysis performance on PDT data of several current LLMS: GPT4 \cite{GPT4}, GPT4o \cite{GPT4o}, Claude Sonnet 3 and 3.5 \cite{Claude}, along with Twitter-Roberta-Base-Sentiment (TRBS) \cite{Roberta}, and Vader (Valence Aware Dictionary and sEntiment Reasoner) \cite{Hutto_Gilbert_2014}. The latter two were chosen for their established effectiveness in sentiment analysis tasks and their differing approaches, which provide a useful comparison to the modern LLMs.  TRBS is a pre-trained bidirectional model fine-tuned on Twitter data for sentiment analysis, based on the RoBERTa architecture, and is a leading transfer learning technique. Vader, a long standing tool in this domain, combines a robust lexicon with heuristic rules for contextual nuance, making it user-friendly and highly accurate. It was designed to perform well on social media text but is effective across other text forms. The selected LLMs (GPT4o and Claude 3.5) represent two of the most advanced LLMs \cite{Krugmann}, and our model  is adaptable for use with other LLMs. 

Two approaches were attempted for running PDT data through the tools to produce PRTG-level scores: 1) having the tool produce sentiments scores for each of the five word/explanation pairs for a respondent and then manually calculating an average (referred to as `Avg5' tests), and 2) having the tool produce one overall sentiment score for all of the word/explanation pairs for a respondent (referred to as `Respondent' tests). From the potential combinations ($tool \bigtimes scoring approach$), we ran the following tests:
\begin{enumerate}
    \item Claude3-Avg5
    \item Claude3.5-Respondent
    \item GPT4-Avg5
    \item GPT4-Respondent
    \item GPT4o-Avg5
    \item GPT4o-Respondent
    \item TRBS-Avg5
    \item TRBS-Respondent
    \item Vader-Avg5
    \item Vader-Respondent
\end{enumerate}

For the LLMs, the associated prompts appear in Table \ref{tab:llm_prompts}. The prompts instruct the LLM to score the PDT data and provide an explanation. Except for Claude3, all LLMs were also asked to provide a confidence level (low, medium, high) for their scoring. These prompts were initially developed and refined through the web interfaces for Claude and GPT. The tests for Claude3-Avg5 and Claude3.5-Respondent were run through the web interface by uploading a CSV file of word/explanation pairs to the web interface, while the other prompts were processed through the GPT API. For GPT4o-Avg5 and GPT4o-Respondent, three runs were conducted and the results were averaged. 

For Avg5 scoring with TRBS and Vader, words and explanations were scored separately and averaged. For respondent-level scoring with TRBS and Vader, the word/explanation pairs were concatenated with a separating period and space ``. '', and the five pairs for the respondent were joined in the same way before being processed by the tool.

\subsection{Method Evaluation}

To assess each tool's performance, well-established metrics for numerical data analysis, such as Pearson Coefficients (PC), Mean Squared Error (MSE) and Mean Absolute Error (MAE) are employed. These metrics provide a comprehensive analysis of sentiment classification, allowing for a thorough evaluation of the system's effectiveness \cite{Bharadwaj}. Additionally, the Wilcoxon statistical test \cite{Rey2011} as well as the paired t-test are used, with the null hypothesis that the mean difference between algorithmic results and the gold standard is zero. The Wilcoxon test is particularly useful when dealing small paired sets of data which do not follow a normal distribution.

\section{Results} \label{results}
\subsection{Overall Results}

Results are shown in Table \ref{tab:sentimentresults}, with tools listed by decreasing order of PC. All strong values are highlighted, and the GPT4o-Respondent approach produced the best statistically significant results, closely followed by Claude3-Avg5. The GPT4o-Avg5 results had strong MAE, MSE, and PC results, but were not statistically matched $(\alpha=.05)$. The other LLMs at the respondent level (Claude3.5 and GPT4) were statistically significant.

\definecolor{lgray}{gray}{0.9}
\definecolor{lgreen}{rgb}{0.56, 0.93, 0.56} 

\begin{table*}[ht]
\centering
\caption{Comparison of Respondent-level Sentiment Analysis Results Ordered by Pearson}
\label{tab:sentimentresults}
\resizebox{\linewidth}{!}{%
\begin{tabular}{|c|>{\centering} p{0.07\linewidth}|>{\centering} p{0.07\linewidth}|>{\centering} p{0.07\linewidth}|>{\centering} p{0.07\linewidth}|>{\centering} p{0.07\linewidth}|>{\centering} p{0.07\linewidth}|>{\centering} p{0.07\linewidth}|>{\centering} p{0.07\linewidth}|>{\centering} p{0.07\linewidth}|>{\centering} p{0.07\linewidth}|c|}
\hline
 \textbf{Bin} & \cellcolor{lgray}\textbf{Gold Standard} & \textbf{GPT4o-Respondent 3RunsAvg} & \textbf{Claude3-Avg5} & \textbf{GPT4o-Avg5 3RunsAvg} & \textbf{Claude3.5-Respondent} & \textbf{GPT4-Respondent} & \textbf{GPT4-Avg5} & \textbf{TRBS-Respondent} & \textbf{Vader-Respondent} & \textbf{TRBS-Avg5} & \textbf{Vader-Avg5} \\ \hline
 $0<=x<0.2$ & \cellcolor{lgray}1 & 2 & 1 & 1 & 1 & 2 & 0 & 3 & 2 & 0 & 0 \\ \hline
 $0.2<=x<0.4$ & \cellcolor{lgray}2 & 1 & 1 & 2 & 1 & 1 & 2 & 0 & 0 & 0 & 0 \\ \hline
 $0.4<=x<0.6$ & \cellcolor{lgray}6 & 5 & 3 & 6 & 2 & 4 & 5 & 4 & 1 & 8 & 35 \\ \hline
 $0.6<=x<0.8$ & \cellcolor{lgray}11 & 14 & 15 & 24 & 13 & 11 & 23 & 9 & 7 & 37 & 15 \\ \hline
 $0.8<=x<=1$ & \cellcolor{lgray}30 & 28 & 30 & 17 & 33 & 32 & 20 & 34 & 40 & 5 & 0 \\ \hline
 \textbf{Totals} & \cellcolor{lgray}50 & 50 & 50 & 50 & 50 & 50 & 50 & 50 & 50 & 50 & 50 \\ \hline\hline
 \textbf{MAE} &  & \cellcolor{lgreen}0.04 & \cellcolor{lgreen}0.03 & \cellcolor{lgreen}0.06 & \cellcolor{lgreen}0.05 & \cellcolor{lgreen}0.06 & \cellcolor{lgreen}0.07 & 0.12 & 0.14 & 0.16 & 0.23 \\ \hline
 \textbf{MSE} &  & \cellcolor{lgreen}0.00 & \cellcolor{lgreen}0.00 & \cellcolor{lgreen}0.00 & \cellcolor{lgreen}0.00 & \cellcolor{lgreen}0.01 & \cellcolor{lgreen}0.01 & \cellcolor{lgreen}0.02 & \cellcolor{lgreen}0.03 & \cellcolor{lgreen}0.04 & 0.07 \\ \hline
 \textbf{Mn} & \cellcolor{lgray}0.10 & 0.09 & 0.10 & 0.15 & 0.15 & 0.05 & 0.31 & 0.04 & 0.03 & 0.54 & 0.48 \\ \hline
 \textbf{Mx} & \cellcolor{lgray}0.94 & 0.97 & 0.94 & 0.87 & 0.95 & 1.00 & 0.88 & 0.99 & 1.00 & 0.91 & 0.74 \\ \hline
 \textbf{Avg} & \cellcolor{lgray}0.76 & \cellcolor{lgreen}0.77 & \cellcolor{lgreen}0.77 & \cellcolor{lgreen}0.71 & \cellcolor{lgreen}0.79 & \cellcolor{lgreen}0.78 & \cellcolor{lgreen}0.72 & 0.82 & 0.88 & 0.68 & 0.58 \\ \hline
 \textbf{SD} & \cellcolor{lgray}0.19 & 0.20 & 0.17 & 0.15 & 0.18 & 0.20 & 0.13 & 0.24 & 0.21 & 0.08 & 0.07 \\ \hline\hline
 \textbf{Pearson} &  & \cellcolor{lgreen}0.97 & \cellcolor{lgreen}0.97 & \cellcolor{lgreen}0.97 & \cellcolor{lgreen}0.95 & \cellcolor{lgreen}0.92 & \cellcolor{lgreen}0.89 & 0.84 & 0.78 & 0.11 & -0.05 \\ \hline
 \textbf{WilcoxonZ} &  & \cellcolor{lgreen}-0.83 & \cellcolor{lgreen}0.46 & 3.43 & \cellcolor{lgreen}1.69 & \cellcolor{lgreen}0.05 & \cellcolor{lgreen}0.22 & 2.79 & 2.79 & 2.65 & 4.51 \\ \hline
 \textbf{t Stat} &  & \cellcolor{lgreen}-1.29 & \cellcolor{lgreen}-1.78 & 6.92 & -3.27 & \cellcolor{lgreen}-1.62 & -2.01 & 3.28 & 6.47 & 2.80 & 6.18 \\ \hline
\end{tabular}
}
\end{table*}

The three GPT4o respondent-level runs were statistically matched, and had between-run PC values of 0.90, 0.92, and 0.92. For the three GPT4o Avg5 runs, the t-stats for run\textsubscript{1} and run\textsubscript{2} means were not statistically matched, but the other runs had between-run t-stats that were significant. These runs had between-run PC values of 0.85, 0.82, and 0.95.

\subsection{LLM Confidence}

All GPT4o runs expressed high confidence for 36 and medium on three PRTGs. On average, GPT4o expressed high confidence on 42 of the 50 PRTGs, and medium confidence on eight. GPT4o did not express low confidence on any PRTG during any run. Claude3.5 expressed high confidence on 31, medium on 8, and low on 11 PRTGs. GPT4 expressed high confidence on 42, and medium on eight PRTGs. 

An example where GPT4o (0.82) and Claude3.5 (0.90) expressed high confidence for a PRTG with positive sentiment (0.90) was explained by GPT4o as: ``The overall sentiment of the words is very positive,  indicating that the project was seen as accessible,  comprehensive,  enjoyable,  useful,  and valuable. The explanations provided further strengthen the positive sentiment by highlighting how the project was organized,  enjoyable,  and educational.''. Claude3.5 explained: ``This respondent's feedback is overwhelmingly positive. They describe the project as `Accessible', `Comprehensive', `Fun', `Useful', and `Valuable'. Their explanations emphasize how enjoyable and beneficial the project was, even calling it one of their favorite projects. The high score reflects the consistently positive feedback with no negative aspects mentioned.''

An example where GPT4o (0.09) and Claude3.5 (0.20) expressed high confidence for a negative sentiment datum was for a PRTG with sentiment of 0.20. One GPT4o explanation was: ``The overall sentiment is overwhelmingly negative due to consistent complaints about complexity,  frustration,  difficulty,  stress,  and the time-consuming nature of the task,  all related to using an IDE that performed poorly.''. Claude3.5 explained ``This respondent's feedback is predominantly negative. They describe the project as `Complex', `Frustrating', `Hard to use', `Stressful', and `Time-consuming'. Their explanations highlight difficulties with the IDE, lack of code comments, and unclear expectations. The low score reflects the consistently negative feedback with no positive aspects mentioned.''

Consistent medium confidence was exampled on a PRTG with sentiment 0.50, where GPT4o averaged 0.54 and Claude3.5 rated 0.50. A GPT4o explanation was: ``The words used,  such as `complex',  `confusing',  and `time-consuming',  have negative connotations, balanced by more neutral or positive words like `comprehensive' and `fun'." However,  the explanations elucidate that while challenges existed,  there was also significant learning and sometimes enjoyment,  contributing to a generally moderate sentiment score.'' Claude3.5 explained: ``This respondent's feedback is mixed. While they found the project `Fun' and `Comprehensive', they also described it as `Complex', `Confusing', and `Time-consuming'. They appreciated the challenge but noted the difficulty in understanding the program initially. The moderate score reflects this balance of positive and negative aspects, with the fun factor slightly outweighing the challenges.''

On average, GPT4o expressed high confidence in its sentiment score for 200 of the 250 word/explanation pairs; medium for 47 pairs, and low confidence for 3 pairs. For 145 pairs, all three runs expressed high confidence in its score, and matched medium confidence for 10 pairs. Only run\textsubscript{2} had any low confidence ratings, and only for two sets of respondent data where the user had not provided an explanation for the words selected. On those cases, the algorithm explained ``Positive but no context provided to adjust.'' 

\subsection{Individual Word-Explanation Pair Results}
\begin{figure}[htbp]
\centerline{\includegraphics[scale=0.38]
{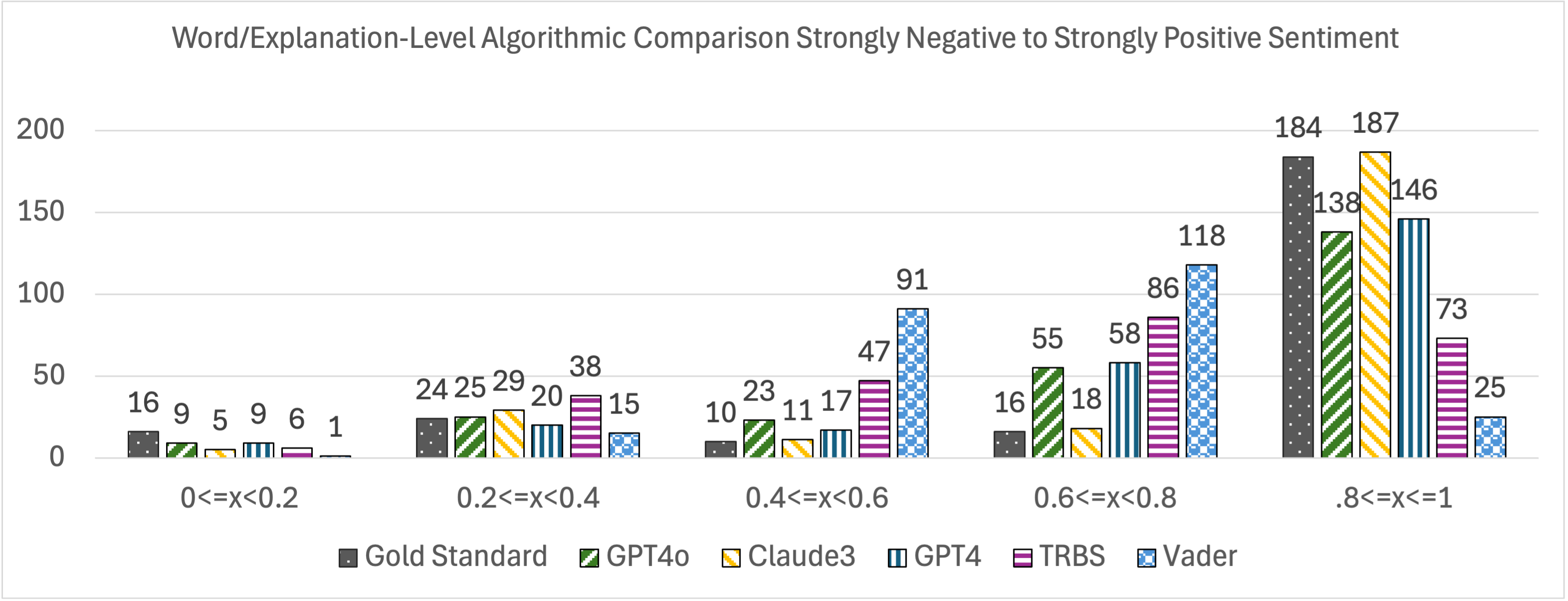}}
 \caption{Word/Explanation Pair Results}
 \label{wordExplainResults}
 \end{figure}

The 250 individual word-explanation pairs were also evaluated through the algorithms (without considering the respondent source), with the hypothesis that the mean difference between algorithmic results and the gold standard for the data is zero. The PC values for the algorithms are Claude3  0.93, GPT4o 0.93, GPT4 0.85, TRBS 0.80, and Vader 0.64. The paired sample t-test and Wilcoxon test failed for all of the algorithms. The results of the algorithmic distribution on the 250 individual word-explanation pairs are shown in Figure \ref{wordExplainResults}. 

\section{Discussion and Future Work} \label{discussion}

The results presented in Table \ref{tab:sentimentresults} demonstrate that the LLMs performed well to quantify implicit user sentiment of PDT data using both respondent-level and Avg5 scoring. Additionally, similar results were found on the CARMA dataset. This research is an initial step toward the creation of a tool for a broad user-base designed to provide rich quantitative sentiment analysis of implicit product desirability, especially in situations where no user rating system exists. Based on the results, either GPT4o or Claude3.5 or a combination of both would serve well as the basis for this tool. A combination could work similar to weighted voting or other bagging-like \cite{bagging} machine learning methods.

The small amount of text in the word/explanation pairs may have made it challenging for the tools to match sentiment, as noted by \citeauthor*{Hartmann} \cite{Hartmann}. However, Claude3 and GPT4 were statistically significant in their match with the gold standard, with GPT4o having good MAE, MSE, PC values using Avg5 scoring, and all were statistically significant with respondent scoring. The results indicate that these LLMs have been able to overcome the challenge of conducting sentiment analysis on short snippets of text. Further work is needed to investigate the impact of data length on LLM performance.

Table \ref{tab:sentimentresults} also provides the sentiment distributions from strongly negative $(x<0.2)$ to strongly positive $(x>=0.80)$, for each tool. These values show the strength of the LLMs, as compared with the other tools, as do the MSE, MAE, and Avg values. A deeper exploration of the approaches for evaluating software desirability used in this paper as a generalized methodology is needed. One area of exploration is the algorithms' ability to match at a given context-level. 

The study also found that the PDT data, especially at the respondent level, provided enough data for LLMs to perform well, without domain-specific training. As a note on the applicability of the PDT as a survey tool for use in studying the effectiveness of gamification applications, it was quick and easy to construct and distribute and has a strong foundation in software product evaluation \cite{Hastings}. Further exploration on other PDT datasets would add verification and support.

The confidence and explanations expressed by the LLMs add value in understanding user sentiment. For example, when GPT4o, GPT4, and Claude3.5 agree in high confidence on their ratings, the authors' confidence in the rating increases, whereas medium or low confidence suggests the potential need for human review. Because the LLMs expressed high confidence a vast majority of the time (GPT4o and GPT4 84\%; Claude3.5 62\%), human review is likely only needed for a few cases. Further exploration is needed for confirmation.

During initial prompt development, Claude3 was provided PDT data in an expansive spreadsheet exported from Qualtrics. Despite the relatively small size (250 rows), it overwhelmed Claude3, preventing it from producing desired behavior. For resolution, data was preprocessed into word/explanation pairs, allowing Claude to focus on the concisely formatted text. In addition, the order of data might be important in some cases. For example, with Avg5 scoring, Claude3 generated consistent base word scores only when pairs were sorted alphabetically by word, effectively grouping word/explanation pairs for the same word together. This observation could be relevant for other LLMs, and further work is needed to determine the impact of PDT data order on scoring.

During initial prompt development, Claude3 demonstrated obvious sentiment analysis capabilities. In contrast, GPT 4 through the web interface struggled to produce consistent behavior. Requesting a sentiment score seemed to appeared to help with GPT4's behavior. Thus, starting with GPT4, a confidence score was also produced. Collectively, a confidence score and an explanation of scoring provided by the LLMs helps in understanding user sentiment. Future work could investigate the quantitative effect on sentiment accuracy of prompting for a confidence score. 

A final observation is that GPT4 through the web interface was cumbersome, and even through the API it had challenges. For any LLM, one off processing of smaller PDT data through the web interface might suffice. To scale, the API is needed.

 \section{Conclusion} \label{conclusion}
 This study adds to a deeper understanding of evaluating user experiences. It explores the use of several LLMs (Claude Sonnet 3 and 3.5, GPT4, and GPT4o) along with TRBS and Vader on a set of PDT data, for providing quantitative numerical zero-shot sentiment analysis of implicit software desirability expressed by users. All LLM tools outperformed the other approaches and were statistically significant in performing as zero-shot sentiment analyzers on the PDT data. The confidence and explanation of confidence provided by the LLMs assist in understanding the user sentiment.

\printbibliography

\end{document}